\documentclass{article} 
\usepackage{nips15submit_e,times}
\usepackage{hyperref}
\usepackage{url}

\usepackage{amssymb}
\usepackage{graphicx}
\usepackage{amsmath,amssymb,multirow,epsfig} 
\usepackage{verbatim}
\usepackage{color}
\usepackage{subcaption}
\usepackage{dsfont}
\title{A Shapley Value Solution to Game Theoretic-based Feature Reduction in False Alarm Detection }

\author{
Fatemeh Afghah, Abolfazl Razi\\
Department of Electrical Eng. \& Computer Sci., 
Northern Arizona University,
Flagstaff, AZ 86011 \\
\texttt{\{fatemeh.afghah, abolfazl.razi\}@nau.edu} \\
\And
Kayvan Najarian \\
Department of Emergency Medicine, University of Michigan,
 Ann Arbor, MI 48109 \\
\texttt{kayvan@med.umich.edu} \\
}

\nipsfinalcopy 

\begin{document}

\maketitle
\begin{abstract} 
False alarm is one of the main concerns in intensive care units and can result in care disruption, sleep deprivation, and insensitivity of care-givers to alarms. Several methods have been proposed to suppress the false alarm rate through improving the quality of physiological signals by filtering, and developing more accurate sensors. However, significant intrinsic correlation among the extracted features limits the performance of most currently available data mining techniques, as they often discard the predictors with low individual impact that may potentially have strong discriminatory power when grouped with others. We propose a model based on coalition game theory that considers the inter-features dependencies in determining the salient predictors in respect to false alarm, which results in improved classification accuracy. The superior performance of this method compared to current methods is shown in simulation results using PhysionNet's MIMIC II database.
\end{abstract}

\section{Introduction} 
Several monitoring and therapeutic devices are utilized in intensive care units (ICUs) to measure vital signs, support or replace impaired or failing organs and administer medications to patients \cite{imhoff2006alarm}. 
Each of these devices might generate optic/acoustic alarms due to patient's physiologic condition, patient movement, motion artifact, malfunction of individual sensors and imperfections in the patient--equipment contact~\cite{philip2009evaluation}.
Many of the alarms (80\% to 99\%~\cite{cvach2012monitor}) could be false and/or clinically insignificant which are not related to patient condition. These alarms could compromise quality and safety of care resulting to many problems such as ``alarm fatigue'' among care--givers as well as the possibility of missing a real event due to care--givers' insensitivity to these unreliable alarms known as ``cry--wolf'' effect. Dealing with false alarms is widely considered the number one hazard imposed by the medical technology and an important concern in ICUs~\cite{cvach2012monitor}. 
Several approaches have been utilized to decrease the number of false alarms as reviewed in ~\cite{cvach2012monitor} and \cite{imhoff2009smart}.

In \cite{li2012signal}, a genetic algorithm-based approach for false alarm reduction is proposed, where the features are extracted from electrocardiogram (ECG), Arterial Blood Pressure (ABP), and Photoplethysmogram (PPG or PLETH) arrhythmia patients. Using a relevance vector machine (RVM) as a classifier, false alarm suppression was reported to be $86.4\%$, $100\%$ and $27.8\%$, respectively for asystole, extreme extreme bradycardia and extreme tachycardia. An automated method for false arrhythmia suppression was proposed in~\cite{behar2013ecg} that is based on quality assessment of normal and abnormal rhythms of ECG signals. Different approaches including k--nearest neighbors (KNN), Naive Bayes, Decision Tree, SVM and multi--layer Perception have been tested on a database from MIMIC II for alarms classification, where features have been extracted from age, sex, Central Venous Pressure (CVP), SpO2, ABP, ECG and Pulmonary
Arterial Pressure (PAP) \cite{baumgartner2012data}. The suppression rate for true alarm detection is between 2.33\% and 17.73\% for 5 alarms and false alarm suppression rate is between 71.73\% to 99.23\%. The aforementioned models considered a number of features/parameters extracted from multiple continuously--measured physiological signals. The main challenge in these multi--parameter approaches is the presence of many parameters / features that individually have low impact on the model performance, which may not be included in the model, while when coupled with other such parameters could significantly improve the accuracy and specificity of the alarm detection algorithms.

Different hybrid feature selection algorithms have been utilized in big data analysis problems to improve the prediction accuracy and reliability through reducing the feature space to a more concise and relevant set of attributes \cite{Saeysreview,Tibshirani,Molina,Peng_Information}. Of the three major approaches for feature selection-- filter-based, wrapper-based \cite{Kohavi, Peng_Information, Huang}, and embedded methods \cite{da06} – the last two are known to be susceptible to overfitting and are computationally intensive \cite{Saeysreview}. The majority of these conventional methods, either only account for the effect of individual features on the target or consider the inter-feature mutual information to obtain higher performance. This often results in discarding the features that are relevant to the target class but are highly correlated to the ones that are already selected. This can significantly degrade the performance of the model in scenarios where a set of features together have a considerable effect on the classifier, while each individual attribute in the set does not \cite{Fan}.

In this paper, we propose a coalition-based game theoretic model for feature selection that accounts for intrinsic inter-features correlation among the predictors across the data sets to improve the accuracy of our model. Coalition game theory has been recently utilized in data analysis problems to improve the performance of feature selection by considering the contribution of the features on classification accuracy when the features are grouped with other features in the data set \cite{Game_EMBC,Sun_cooperative,Game_BHI,Cohen_journal}. In \cite{Game_BHI}, we proposed a coalition-based game theoretic feature selection method to determine the salient features over a heterogeneous data set to predict the hemorrhage severity, where the features are modeled as the game players. The importance of each feature in the game is measured by Shapley value defined as its contribution in improving the classification accuracy considering all possible coalitions of the features. In \cite{Game_EMBC}, we developed a network-based coalition game theoretic framework to discover the most informative gene subnetworks in predicting ovarian cancer by integrating gene expression profiling of cancer tissues with protein-protein interaction (PPI) networks. This model considered the genes as the game players and develops pathways emerging from a seed gene set in PPI network by traversing the network to discover the most informative pathways associated with a desired outcome computing the Shapely value of the players. 

Here we describe a recently proposed coalition-based game theoretic model in \cite{Game_ICSH} to suppress the false alarm using three signals of ECG, PLETH and ABP from Physionet's MIMIC II database. The impact of each feature in the game in interaction with other features when they form a coalition is measured by Multi--perturbation Shapely for coalitions of size 4, which results in significant accuracy improvement comparing to other feature selection techniques including Chi--square, Gain Ratio, Relief and Info Gain methods. 

\section{Signal Processing and Feature Extraction} 
In this study, five types of life threatening arrhythmias including asystole, extreme bradycardia, extreme tachycardia, ventricular tachycardia, and ventricular flutter/fibrillation are considered. Three main signals; ECG, ABP, and PLETH are used as the inputs of our proposed model. In the first stage (i.e. signal analysis) wavelet coefficients of each signal at different levels of decomposition are calculated using a discrete wavelet transform (DWT) on the 1--D input signals. 
At each level of decomposition process, DWT decomposes the signals into approximate and detail coefficients. Approximation set is obtained by applying a high--pass filter at low scales and detail coefficients are computed by applying a low--pass filter at high scales. We used Daubechies 8 (db8) for ECG signal as there is a good match between the shape of ECG signal and this wavelet. Daubechies 4 is used for PLETH and ABP signals for the same reason.  
Each of the three aforementioned signals are decomposed into $6$ levels by convolving the high-pass and low--pass filters. Feeding all these wavelet coefficients as features into the classification algorithm is not efficient and may significantly decrease the generalization property of the trained model due to over--fitting. Therefore, we reduce the number of features by extracting 20 representative statistical and information--theoretic properties of the wavelet vectors.

\section{Coalition-based Game-theoretic Feature Selection} 
Cooperative game theory has been recently utilized in feature selection algorithms \cite{Sun_cooperative,Game_BHI,Game_EMBC,Cohen_journal}. In these games, the players cooperate with each other by forming various sub--groups called \emph{coalitions}. These games are defined based on exhaustive scenarios that players may form a group and how the total shared payoff is divided among the members. A transferable utility coalition (TU--coalition) game with $n$ players can be defined by $(\mathcal{N},v)$, where $\mathcal{N}$ denotes the set of players, $\mathcal{N}=\{1,2,...,n\}$, and \emph{
characteristic function}, $v$ is a real--valued function defined on the set of all coalitions, $v:2^\mathcal{N} \rightarrow \mathbb{R}$. 
For a coalition $\mathcal{S}$, $ \mathcal{S} \subseteq \mathcal{N}$, the characteristic function, $v(\mathcal{S})$ represents the total payoff that can be gained by the members of this coalition, and satisfies the following conditions, i) characteristic function of an empty coalition $\phi$ is zero, $v(\phi)=0$, and ii) if $\mathcal{S}_i$ and $\mathcal{S}_j$,  ($\mathcal{S}_i, \mathcal{S}_j\subseteq \mathcal{N}$) are two disjoint coalitions, the characteristic function of their union has super--additivity property, meaning that $v(\mathcal{S}_i \cup \mathcal{S}_j)\geq v(\mathcal{S}_i)+v(\mathcal{S}_j)$. 

Here, we model the features as the players of the game, and the characteristic function of a coalition, $v$ is measured by contribution of its members (features) to the performance of the classifier (e.g. success rate in supervised learning). 
Different possible grouping of the features are examined to recognize the optimal coalition. 
The contribution of feature $i$ in classification accuracy when it joins a coalition $\mathcal{S}$ is defined by \emph{marginal importance} as follows 
\begin{align}
\Delta_i(\mathcal{S})=v(\mathcal{S} \cup \{i\})-v(\mathcal{S})
\end{align}

A solution of a coalition game is determined by how the coalition of players can be formed and how the total payoff of a coalition is divided among the members. 
Let's define the value function, $\gamma$ that assigns an $n$--tuple of real numbers, $\gamma(v)=(\gamma_1(v), \gamma_2(v), ..., \gamma_n(v))$ to each possible characteristic function, in which $\gamma_i(v)$ measures the value of player $i$ in the game with characteristic function $v$. \emph{Shapley value} can be utilized as a fair unique solution of the coalition game \cite{Shapley}. 
The Shapley value of player $i$ is defined as the weighted mean of its marginal importance over all possible subsets of the players. \vspace{-4pt}
\begin{align} \label{eq:sh1} 
\gamma_i(v)=\frac{1}{n!} \sum_{\pi \in {\Pi}} \Delta_i(\mathcal{S}_i(\pi)),
\end{align}
where $\Pi$ is the set of all $n!$ permutations over $\mathcal{N}$ and $\mathcal{S}_i(\pi)$ is the set of features (players) preceding player $i$ in permutation $\pi$.
Since in feature selection, the order of features in a coalition does not change the value of coalition, the calculations in (\ref{eq:sh1}), can be further simplified by excluding the permutation of coalitions in the average:  \vspace{-4pt}
\begin{align} \label{eq:sh2}
\gamma_i(\mathcal{N},v)=\frac{1}{n!}\sum_{\mathcal{S}\subseteq \mathcal{N}/i}\Delta_i(\mathcal{S})|\mathcal{S}|_i(n-|\mathcal{S}|-1))!,
\end{align}
where ${\mathcal{S}\subseteq \mathcal{N}/i}$ presents the coalitions that player $i$ does not belong to. It is equivalent to the weighted average of coalitions, where the weight of each coalition is the number of its all possible permutations.
As shown in (\ref{eq:sh1}) and (\ref{eq:sh2}), the Shapely value solution accounts for all possible coalitions that can be formed by the players \cite{Shapley}. Since in false alarm detection problem, the data set includes a large number of features, thereby calculating the Shapley value would be computationally intractable. 
Therefore, we utilize the Multi--perturbation Shapley value measurement with coalition sizes up to $L$ rather than the original Shapely value, which is determined using an unbiased estimator based on Shapley value \cite {Keinan,Kaufman}.

In our proposed algorithm, at each round, the features are randomly divided into groups of size $L$. 
Then, we calculate the corresponding Multi-perturbation Shapely value of feature $i$ inside its group, $\gamma'_i(v)$ considering all possible coalitions of size $1 \leq l \leq L$. This is equivalent to randomly sampling from uniformly distributed feature $i$, $\gamma'_i(v)$ is calculated as follows.
\begin{align} \label{MSA}
\gamma'_i(v)=\frac{1}{|\Pi_L|} \sum_{\pi \in {\Pi_L}} \Delta_i(S_i(\pi)),
\end{align}
where $\Pi_L$ denotes the sampled permutation on sub--groups of features of size $L$. There is an essential trade--off to set $L$ in the proposed method. Large $L$ values consider higher order relations, while increasing the complexity of finding Multi--perturbation Shapely value at each subgroup. We conjecture that the optimum value of $L$ for our datasets taking into account various factors such as the nature of data, number of features, and the inter-feature dependence is in the range of $3$ to $6$. This is confirmed by simulation results in section \ref{numerical}.

\section{Numerical Analysis}  \label{numerical}
For this study, we used a publicly available PhysionNet's MIMIC II databaset~\cite{PhysioNet}, where measurement for three vital signals ECG, PLETH, and APB are provided for $219$ patients and each subject is labeled as \textit{true}, \textit{false}, or {impossible to tell}. We first apply six-level wavelet decomposition to obtain time-frequency information at different resolutions. Therefore, each sample is represented by $18$ vectors of wavelet coefficients. Subsequently, $20$ statistical and information-theoretic features are extracted from each of the vectors, resulting in total of $180$ features.
Experimental results are provided in this section for the proposed alarm validation method as well as other state--of--the--art explicit feature selection methods including Chi--square, Gain Ratio, Relief and Info Gain methods. 
The numerical results are obtained utilizing the proposed coalition--game theoretic method where the multi--perturbation Shapley value is calculated for coalitions' size up to 4, $L=4$. 
The alarm typing rate for all feature selection methods are evaluated in combination with Bayes Net classification as a representative classifier. 
In all simulations, the 30 most informative features are selected to compare the performance of different feature selection techniques.

\begin{figure}[ht!]
	\centering
	\begin{subfigure}[t]{0.5\textwidth}
		\centering
		\includegraphics[width=1.1 \linewidth]{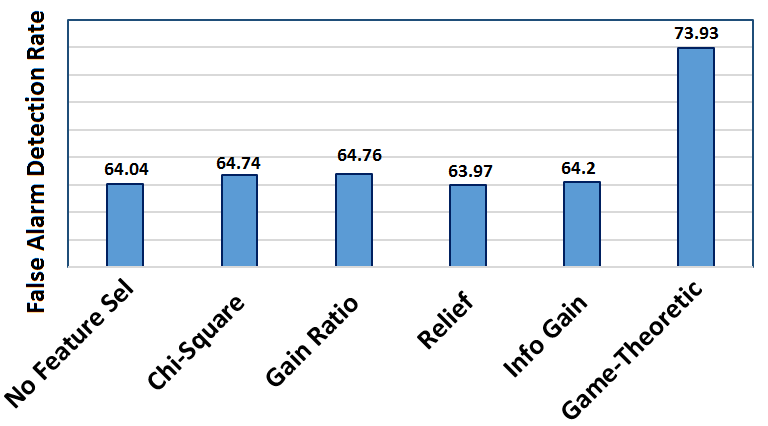}
		\caption{False alarm rate detection for the first 30 features using different feature selection methods with Bayes Net classification.}
	\label{fig:1}
	\end{subfigure}%
	~
	\begin{subfigure}[t]{0.5\textwidth}
		\centering
		\includegraphics[width=0.8 \linewidth]{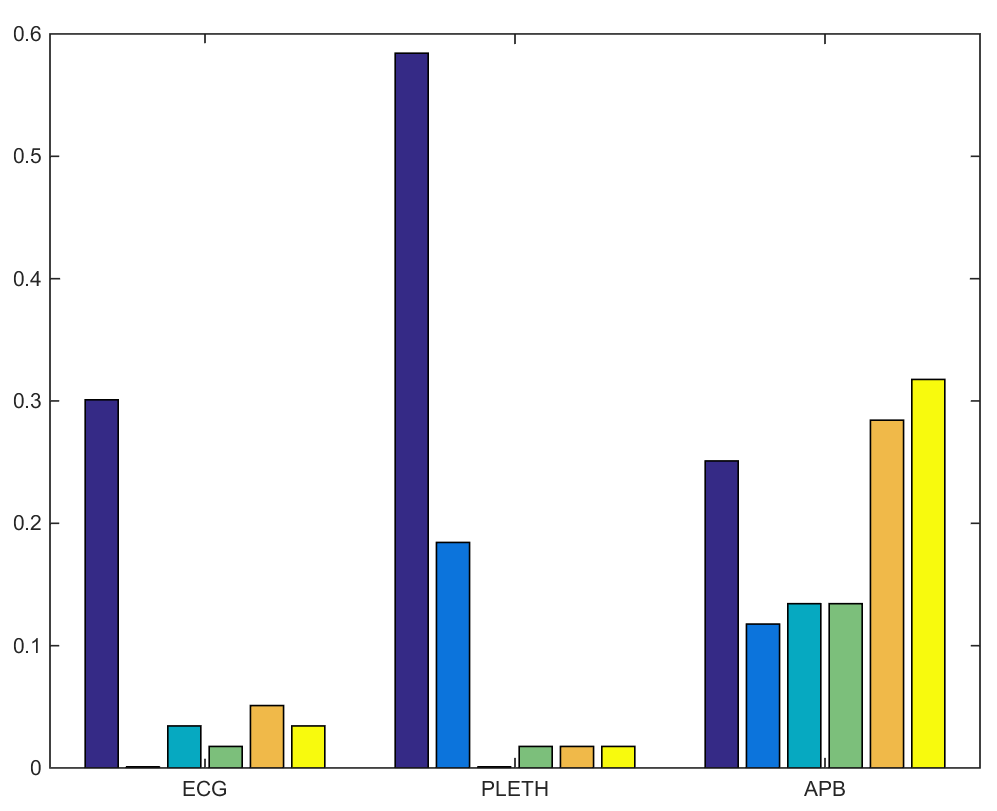}
		\caption{Average appearance of the six levels of wavelet decomposed vectors for ECG, PLETH and APB signals}
\label{fig:2}
	\end{subfigure}
	
\end{figure}

The comparison results in Fig. \ref{fig:1} suggest a considerable improvement for the proposed method in discarding the false alarms compared to the competitor methods. The alarm typing success rate for the proposed method is about $75\%$ meaning that only $25\%$ of alarms are deemed false, whereas the false alarm report rate for the best competitor method (Gain Ratio) is at least $100-68.88\% \approx 31\%$. The improvement is due to potential synergy impact of coalitions among features which is overlooked or not directly addressed in other methods. The proposed method outperforms the case of incorporating all wavelet coefficients (represented by None in Fig. \ref{fig:1}) due to eliminating the irrelevant features.
It is notable that the promising rate of $75\%$ is obtained using only 30 statistical features for any subject, which significantly reduces the risk of over--fitting compared to using all $18000$ wavelet coefficients for each signal. The average appearance of signals are depicted in Fig. \ref{fig:2}. It is clear from these results that the first wavelet decomposition level of \textit{ECG} and \textit{PLETH} signals play significantly higher roles in the alarm validation. Indeed, the collective contribution of levels 2 to 6 are less than the contribution of level 1 solely. However, all levels of signal \textit{APB signal} contribute almost equally for alarm recognition.

%
%

\newpage
\small{
\bibliographystyle{IEEEtran}
\bibliography{IEEEabrv,Ref}}

%
%
%
%

\end{document}